\documentclass[11pt]{article}

\usepackage{longtable}
\usepackage{booktabs}
\usepackage{array}
\usepackage{amsmath}
\usepackage{amsmath}
\usepackage{amssymb}
\usepackage{comment}
\usepackage{booktabs}
\usepackage{tabularx}
\usepackage{acl}
\usepackage{graphicx}
\usepackage{latexsym}
\usepackage[T1]{fontenc}
\usepackage{tgheros}
\usepackage{tgcursor}

\usepackage[T1]{fontenc}

\usepackage[utf8]{inputenc}

\usepackage{microtype}

\title{Measuring the practice of shared-decision making (OPTION12): An Investigation into Open-sourced Smaller LLMs (OS-sLLMs) for Better Privacy and Sustainability}

\author{ Tamara Wit$^{1\dagger}$, Lifeng Han$^{1,2\dagger}$, Carly Heipon$^1$, David Lindevelt$^2$ \\
\textbf{Anne Stiggelbout$^1$, Suzan Verberne$^2$}
 \\ On behalf of the 4D PICTURE consortium \\
\texttt{BDS, Leiden University Medical Centre, NL} \\
 \texttt{The Leiden Institute of Advanced Computer Science, LU, NL} \\
$^{\dagger}$ co-first, corresponding: \{t.wit, l.han\} @ lumc.nl  \\
 \\
 \\
}
\begin{document}
\maketitle
\section*{Accepted Abstract}
\textbf{Background}: Shared decision-making (SDM) is important in clinical consultations in which patients discuss and decide on treatment options with clinicians. This SDM process is often coded with the OPTION12 instrument, an observer-based tool, consisting of 12 items, rated on a 5-point Likert scale. This coding is conducted by human coders. Coding is time-intensive and frequently accompanied by disagreement between coders. We explore the capability of open-source privacy-preserving smaller LLMs (OS-sLLMs) to perform the coding task, potentially automating the process, with humans in the loop. 
\textbf{Methods}: 26 transcripts of Dutch melanoma patients consultations with clinicians were double-coded by two coders. Two human coders resolved their disagreements after independent coding. To evaluate OPTION12 coding using OS-sLLMs, the consultation data was divided into development and testing sets. We designed the complete investigation framework (see Figure \ref{fig:llm4sdm-diag}). It includes 1) a pilot study of the development set (11 interviews) for prompt-refinement and sLLM selection. 2) deployment of the fine-tuned prompts and best performing sLLM (judge-sLLM) augmented with few-shot examples from the development set on the testing set (15 interviews) and asking the judge-sLLM to resolve the disagreement on other OS-sLLMs’ scores.
The rationale for judge-llm: it behaves like a third-party human to resolve the disagreements of annotators. Instead of using humans, a judge-LLM is consulted to resolve disagreement. This is designed to mimic how human annotators resolve disagreements, by discussing until consensus is achieved. We plan to investigate alternative model-merging methods for the future.
First, for prompt refinement, we use chain-of-thoughts (CoTs), LLM-assisted prompting, human-in-the-loop with sample output (few-shots) feedback.
For OS-sLLMs, we use both 1) general domain models Llama, Gemma, and Mistral7b, and 2) medical domain models Meditron and Medllama. 
Second, for judge-sLLM selection, on the system-level, we measure the overall correlation of each sLLM and human coding using Spearman and Pearson correlation scores. On the segment-level, we also look into the most agreed-upon and disagreed-upon items . We will perform both qualitative and quantitative analysis by discussing the evaluation scores and categorise the OS-sLLMs’ behaviours with examples.
Finally, we deploy the OS-sLLMs on the testing set and use the judge-sLLM to resolve disagreements.

\textbf{Preliminary} results: Five OS-sLLMs show the following findings: 
\begin{itemize}
    \item Three general domain OS-sLLMs perform better than the two medical domain ones, which both generate hallucinations and do not follow prompts precisely, indicating further developments is needed for medical OS-sLLMs.
    \item Mistral7b outperformed the other two Gemma3:12b and Llama3.1:8b by 4 consensus with human coding, vs 3 items.
    \item The overall correlation with human coding on these 12 items is (0.83, 0.80, 0.64) using Pearson correlation, and (0.81, 0.78, 0.61) using Spearman rank correlation from the three models (gemma3:12b, llama3.1:8b, mistral7b).
    \item For the items for which OS-sLLMs agree with human coders, OS-sLLMs can generate the same sentences as humans in some cases, but in other cases, they generate better quotes than humans.
\end{itemize}

\textbf{Impacts}: This study reports the first research findings using OS-sLLMs on scoring SDM with the OPION12 in Dutch melanoma patients’ consultation transcripts. The performance of such OS-sLLMs is promising and valuable to our task on measuring patient involvement. They have agreements with human coders on certain items and we are looking into the disagreement to see if we can have different inputs from OS-sLLMs, or if we can fine-tune OS-sLLMs to achieve human level performances on the rest of the items. In the long-term view, we expect fine-tuned OS-sLLMs can achieve human level performance and be deployed for coding tasks with humans in the loop for verification and quality control.
\footnote{{extended work based on accepted Abstract at ISDM2026 \url{https://sites.dartmouth.edu/isdm2026}}}

\newpage
\section*{Extended work from Conf Abstract}
\section{Introduction}


Shared Decision Making (SDM) is very important for healthcare since it encourages patients and their family members to get involved in the decision-making process of their treatments \cite{stiggelbout2015shared}.
There are regulations for SDM guidelines, such as OPTION12; however, there is the coding burden from human annotators and such manual work.
 
There are articles on using LLMs for medical decision making collaborations, but rarely on predicting the level of SDM in practice, with the only one we found on 
breast cancer only \cite{selvaraj2025automating}.
It is on a different coding scheme (OPTION5), and it uses commercial GPT models (GPT-4o, Gemini-1.50pro) that are very large, costly, and have potential privacy/ethical concerns regarding patient/medical data usage.

To promote sustainable development, smaller-sized LLMs need to be tested for such a task (e.g. instead of Llama 405b,
Llama 70b); to protect privacy, open-sourced, local LLMs need to be researched and deployed.
There is a research Gap:
how to measure OPTION12 SDM performance using open-source models for better privacy. In our case, we have Dutch melanoma patient data.



OUR research objective: 
How open-sourced LLMs perform in predicting the SDM implementation levels in the clinic and if their performance can be improved with fine-tuning using human-annotated data; furthermore, if LLMs can be integrated into this task to save manual annotation cost.

Research Questions:
\textbf{RQ-1a}: if state of the art open-sourced smaller LLMs (OS-sLLMs) can perform human-like annotation regarding the levels of SDM implementations in clinical practice? 
\textbf{RQ-1b}: 1) if LLMs perform differently with human annotators, do they give a different point of view, or just nonsense? 2) if they score the same as humans, do they pick the same quotes?
\textbf{RQ-1c}: which of the 12 items in OPTION12 do LLMs tend to perform better/worse? what are the overall correlation levels with human coding?
\textbf{RQ-1d}: do general domain LLM perform differently vs medical domain LLMs?

\textbf{RQ-2}: can fine-tuning LLMs (or example-based learning) using human annotated data help them on performing this specific task on SDM implementation levels from Scheme-OPTION12?

\textbf{RQ-3}: can we use LLMs to conduct this annotation task if it can perform fairly, e.g. LLMs + human in the loop, so as to save cost?

Highlighted contributions of this work are:
\begin{itemize}
    \item First investigation of open-source small LLMs for OPTION12 coding.
    \item Comparison of: general-domain OS-sLLMs and medical-domain OS-sLLMs for SDM assessment.
    \item Discovery that: medical models hallucinate more; general models perform better; despite domain specialization (a surprising result).
    \item Identification of systematic SDM-coding error categories (taxonomy): temporality failures, role confusion, evidence hallucination, score-evidence mismatch, option misunderstanding. 
    \item Proposal of the Judge-LLM consensus framework. 
\end{itemize}

\section{Related Work}
We describe related and background work in SDM and OPTION scale, regional or domain-specific SDM, and LLM investigation for SDM.

\subsection{SDM and the OPTION scale}
The early works on developing the OPTION scheme for SDM include \cite{elwyn2005shared}, where the authors designed the scales to measure patient involvement in the SDM procedure
\cite{elwyn2003shared}.

Later, there are inspired academic discussions and adaptations of such designs into different sub-fields in healthcare, for instance,
SDM in Psychometric instrument \cite{simon2007measuring} and
psychometric study of OPTION5 \cite{barr2015psychometric}; dyatic OPTION for perceptions of SDM \cite{melbourne2010developing}; 
comparing OPTION and informed SDM instruments \cite{weiss2008measuring};
discussing methodological issues of the OPTION scale \cite{nicolai2012option};
OPTION - convergent validity \cite{scholl2015comparing};
OPTION5 vs OPTION12 \cite{stubenrouch2016option5} comparison.
Experimental results from earlier works show that the OPTION5 scores can be consistently higher than the OPTION12 ones. We think this can be caused by the fact that  OPTION12 includes more aspects to measure.
In our current research project, we use melanoma patients' consultation data on the OPTION12 scales.
More SDM overview works can be found at \cite{scholl2011measurement,couet2015assessments,gartner2018quality,ahmad2020shared}.

\subsection{Region and Domain-Specific SDMs}

Another line of work related to ours is the regional or domain-specific SDMs, since we are focusing on Dutch Melanoma patients.
Regarding regional SDM, such works include the study of the reliability of the OPTION score in Italy \cite{goss2007shared}, 
SDM in Canada \cite{elwyn2013using}.
Focusing on specific patients, there are
SDM in outpatient chronic care \cite{norful2020instruments},
SDM in cardiovascular care \cite{sepucha2014measuring}, as well as
OPTION score for resident patient consultations in family medicine \cite{pellerin2011toward}.


\subsection{LLMs on SDM}



LLMs have been suggested by researchers to assist the SDM process by offering lay language adaptation of medical evaluations and evidence as a collaborator role with the clinician \cite{elwyn2024meet}.

\cite{selvaraj2025automating} present the first empirical study investigating whether large language models (LLMs) can automate the observer-based assessment of shared decision making (SDM) using the Observer OPTION-5 instrument. The study evaluates commercial LLMs from the GPT, Gemini, and LLaMA families on 287 anonymized transcripts of clinician-patient consultations involving women considering surgery for early-stage breast cancer. Human-rated OPTION-5 scores from a randomized controlled trial were used as the reference standard. The authors tested multiple prompting strategies and assessed agreement between LLM-generated and human-generated scores using correlation analysis.

The study found that GPT-4o and Gemini-1.5-Pro achieved moderate correlations with human ratings (Pearson correlation approximately (r $\approx$ 0.6)), corresponding to roughly 75–80\% of the agreement observed between human raters themselves ((r = 0.77)). Performance improved when prompts included detailed item descriptions and examples, suggesting that prompt engineering substantially affects LLM performance on SDM coding tasks. Additionally, the models were able to distinguish between consultations with high versus low SDM implementation levels, indicating potential utility for automated quality assessment.

This work differs from our work in several ways. First, they focus on the Observer OPTION-5 instrument, whereas our study investigates the more fine-grained OPTION-12 coding scheme. Second, their work uses large commercial models (GPT-4o and Gemini-1.5-Pro), while our study evaluates privacy-preserving, open-sourced, smaller LLMs (OS-sLLMs) that can be deployed locally. Third, their dataset concerns breast cancer surgery consultations in English, whereas our work focuses on Dutch melanoma consultations. Finally, while they primarily assess model validity through correlations with human ratings, our study additionally investigates model disagreement patterns, qualitative error categories, and the potential role of judge-LLMs in resolving scoring disagreements.



\section{Methodology}

\begin{figure*}[t]
  \includegraphics[width=1\textwidth]{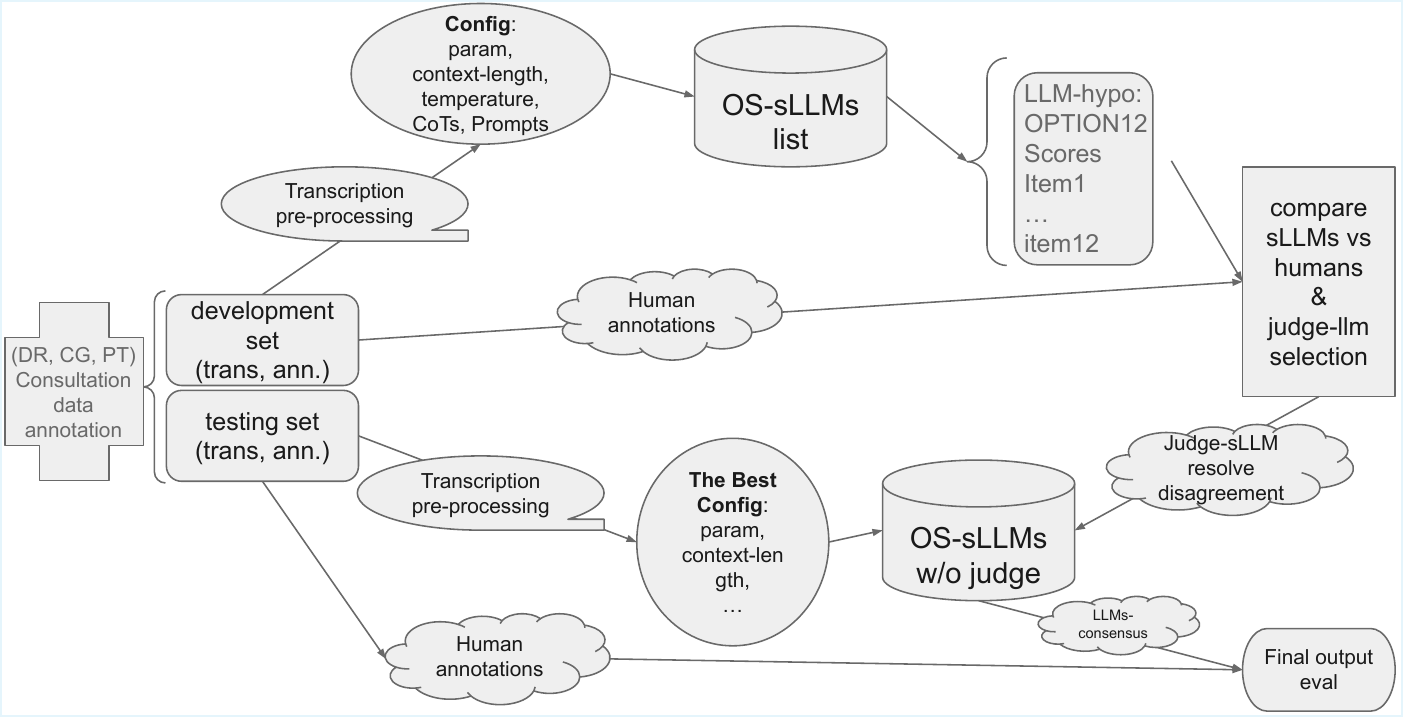}
  \caption{An overview of \textsc{LLM4SDM} framework: upper layer for development phase and lower layer for model testing.
  }
  \label{fig:llm4sdm-diag}
\end{figure*}

The methodology framework is presented in Figure \ref{fig:llm4sdm-diag}, where the upper half is the developmental phase and the lower half is the deployment phase using tuned LLMs.
From left to right, we start with transcripts from melanoma patient where a decision took place. These transcripts have been coded by two researchers (CH/TW). All transcripts were individually coded, consensus meetings were held to discuss potential disagreements until consensus was achieved. The data includes three roles DR, CG, and PT.
We split the coded data into development (pilot study) and testing sets. 
For the \textbf{development} phase, transcription pre-processing was carried out for NLP purpose. 
For model configuration, we explored parameter optimization, context-length and temperature adjustment, prompting with Chain-of-Thoughts (CoTs).
We select a list of open-sourced smaller-sized LLMs (OS-sLLMs) covering both general and medical models.
Using the selected OS-sLLMs and our model configuration, we produce the automatic system outputs (i.e., annotations and justifications) of SDM scores from each file in the pilot study. We compare the automatic annotations to the human annotations both quantitatively using correlation scores and qualitatively looking into errors or disagreements.  
At the end of the pilot study phase, we 1) conduct the categorization of LLM-related issues on this task; 2) select the best performing OS-sLLMs as a judge (Judge-sLLM) for the next phase.  

In the \textbf{testing} phase, as in the lower part of the methodology figure (Figure \ref{fig:llm4sdm-diag}) from both left and right to the middle, we first carry out the same pre-processing for the documents in the test set as the same procedure of the pilot study. Then, we deploy the parameter-optimized OS-sLLMs w/o judge on the test set through few-shot (example-based) and instruction-based fine-tuning to get the automatic annotations. Thirdly, we use Judge-sLLM to resolve the consensus from the outputs produced by other OS-sLLMs to mimic how human annotators would carry out this work.
Lastly, the final output from consensus-ed LLMs will be compared with human annotations and evaluated on the test set.


\noindent \textbf{Judge-LLM Consensus Strategy}
Human OPTION12 coding typically involves independent annotation followed by consensus discussion. Let \(h_1\) and \(h_2\) denote two human annotators. Their independent annotations are reconciled into a final human consensus label:

\[
(h_1, h_2) \rightarrow y_{\mathrm{human\ consensus}}.
\]

We adapt this principle to LLM-based annotation. Let \(m_1, m_2, \ldots, m_k\) denote the outputs of multiple OS-sLLMs for the same consultation and OPTION12 item. A selected Judge-LLM \(J\) receives these candidate annotations and produces a consensus output:

\[
J(m_1, m_2, \ldots, m_k) \rightarrow y_{\mathrm{LLM\ consensus}}.
\]

In this setting, each candidate annotation includes the predicted OPTION12 score, supporting evidence, and justification. The Judge-LLM is therefore not only asked to select or reconcile scores, but also to assess whether the provided evidence is grounded in the transcript and consistent with the OPTION12 item definition.

\section{Experiments on Development Set}

\subsection{Data and Models}

We have 26 transcriptions of Shared Decision Making consultations of \textbf{melanoma} patients, which are \textbf{double-coded} by two researchers in the healthcare sector,
and agreed on consensus with the OPTION 12 coding scheme (Observing patients' involvement in decision making). 
They have numerical values (0-to-4) for a list of 12 categories. 
We applied the Refined Manual Scheme (RMS), a task-oriented scheme, revised from the existing validated coding scheme (on the public website). \footnote{\url{https://each.international/reachresources/option-12/}}
The coders carried out coding independently and resolved the initially disagreed-upon coded values by discussion. The final gold reference coded values are the human-agreed ones.
We split this data set into development/validation and test sets (11, 15).

To investigate both general and medical domain OS-LLMs, we used the following comparable-sized lightweight models: 1) general domain pre-trained Gemma3:12b, Llama3.1:8b, and Mistral7b, and 2) medical domain finetuned models MedLlama2:7b and Medtron7b.




\subsection{Evaluation Metrics}

We define several metrics to evaluate the agreement between LLM-generated annotations and human annotations under the OPTION12 scheme.

\paragraph{Total Consensus.}
The total number of agreements between a model and human annotators across all files and items is defined as:\begin{equation}
\text{TotalConsensus} = \sum_{f=1}^{N} \sum_{i=1}^{12} C_{f,i}
\end{equation}

\paragraph{Mean Consensus per File.}
The average number of agreed items per file is computed as:
\begin{equation}
\text{MeanConsensus} = \frac{1}{N} \sum_{f=1}^{N} \sum_{i=1}^{12} C_{f,i}
\end{equation}

\paragraph{Normalized Mean Consensus.}
To obtain a percentage-based metric, we normalize the mean consensus by the total number of items:
\begin{equation}
\text{MeanConsensus}_{\text{norm}} = \frac{1}{12N} \sum_{f=1}^{N} \sum_{i=1}^{12} C_{f,i} \times 100
\end{equation}

\paragraph{Per-item Agreement Ratio.}
For each model $m$ and item $i$, we compute the agreement ratio across all files:
\begin{equation}
\text{Ratio}_{m,i} = \frac{C_{m,i}}{N}, \quad 
C_{m,i} = \sum_{f=1}^{N} \mathbb{1}(\hat{y}_{m,f,i} = y_{f,i})
\end{equation}

where $C_{f,i} \in \{0,1\}$ indicates whether the model's prediction matches the human annotation for file $f$ and item $i$, $C_{m,i}$ is the total number of correct predictions made by model $m$ for item $i$, $N$ is the number of files (patients), and $\mathbb{1}(\cdot)$ is an indicator function that equals 1 when the condition holds and 0 otherwise.

In practice, we report $\text{Ratio}_{m,i}$ as percentages.

We normalize total agreement counts per item by the total number of model-file combinations ($M \times N$) to obtain percentage agreement rates.

\begin{equation}
\small 
\text{TotalCorrectRate}_{i} = \frac{100}{MN} \sum_{m=1}^{M} \sum_{f=1}^{N} \mathbb{1}(\hat{y}_{m,f,i} = y_{f,i})
\end{equation}

\subsection{Pilot Study Outcomes}

% ...

To answer our \textbf{RQ-1d}: Do medical and general domain LLMs perform differently? Yes, the medical LLMs have more hallucinations.

\subsubsection{Hallucinations of Medical LLMs}

\begin{figure}[t]
  \includegraphics[width=.45\textwidth]{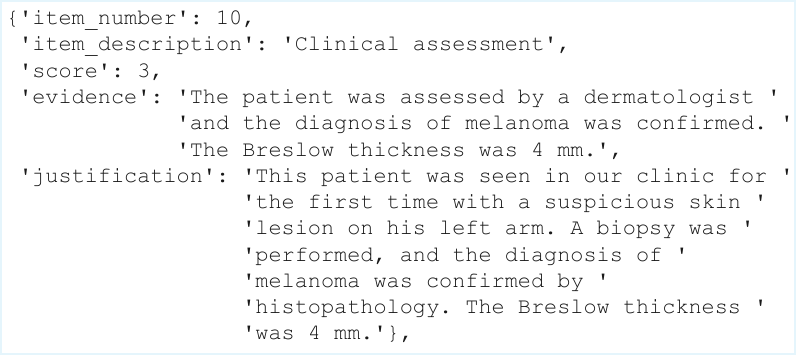}
  \caption{Meditron Model Hallucination Example - repeated evidence and justification across predicted OPTIOIN12 items using the same score 3.
  }
  \label{fig:hallucination-all-items_meditron_7b.130002_NL}
\end{figure}

\begin{figure}[t]
  \includegraphics[width=.45\textwidth]{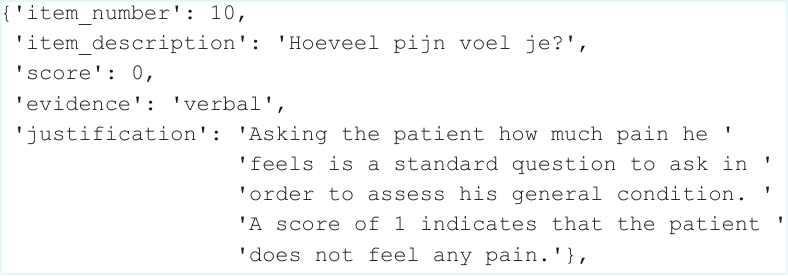}
  \caption{Meditron Model Hallucination Example - giving some score 0 for all OPTION12 items but using similar text to justify score 1 instead.
  }
  \label{fig:2nd_run_hallucination-all-items_meditron_7b.130002_NL}
\end{figure}

\begin{figure}[t]
  \includegraphics[width=.45\textwidth]{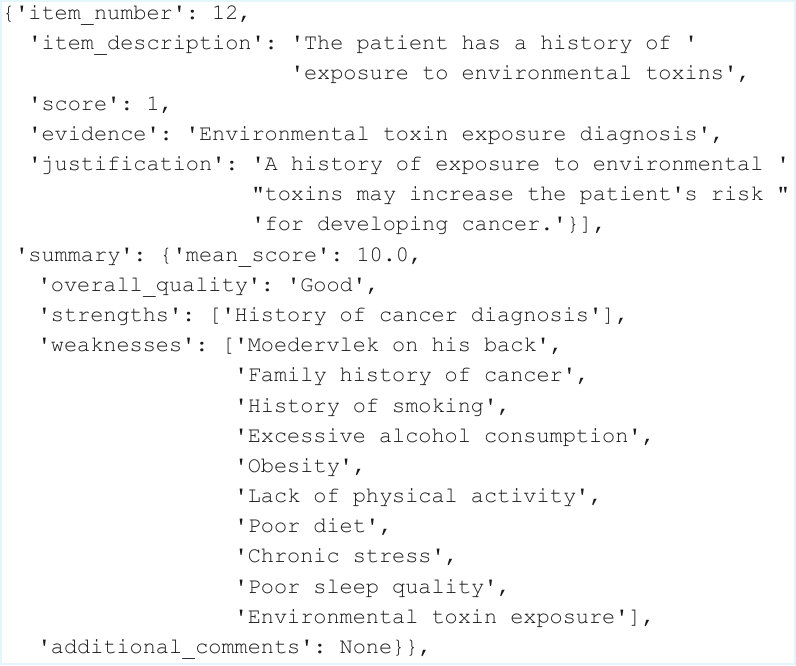}
  \caption{MedLlama Model Hallucination Example - giving score 1 for all OPTION12 items but mean score 10.
  }
  \label{fig:hallucination-all-items_score1_medllama2_7b.110019_NL}
\end{figure}

The two medical OS-sLLMs both generated hallucinated outputs that we have to give up on these tested models: MedLlama and Meditron.

Example of Meditron7b model hallucination behavior on output is show via patient file ``130002'' in Figure \ref{fig:hallucination-all-items_meditron_7b.130002_NL}, where in the first run, the model generated score 3 for its outputs only covering partial OPTION12 items (Item 10, 11, and 12) using the same repeated evidence and justification across these items. However, in the 2nd run, after we fix its missing items in the output by more strict instruction, it generated score 0 for all OPTION12 items (Item 1 to 12) but using similar evidence and justification phases to justify score 1, as in Figure \ref{fig:2nd_run_hallucination-all-items_meditron_7b.130002_NL}.





MedLlama also produced hallucination in the initial deployment when it generated score 1 for all OPTION12 items and even gave a mean score 10 very wrongly by averaging individual scores, as in Figure \ref{fig:hallucination-all-items_score1_medllama2_7b.110019_NL} from example patient file ``110019''. However, it did give different justifications and evidences across OPTION12 items.


\subsubsection{Metric scores from general LLMs}

Out of 11 development files, 3 files did not work out due to formatting, and 1 did not receive human code because it was not suitable for SDM. 
Adjustment:
So, in total, we have 7 patient files (representing 7 clinician-patient consultations) successfully run through the development phase, as in Table \ref{tab:dev2seven}.
The total number of model-file combinations is 
$M \times N = 3 \times 7 = 21$.

\begin{table}[ht]
\small
\centering
\caption{Consultation selection for the development phase.}
\label{tab:consultation_selection}
\label{tab:dev2seven}
\begin{tabular}{lr}
\hline
Step & N \\
\hline
Total consultation transcripts & 26 \\
Development set & 11 \\
\hspace{3mm}Excluded: preprocessing failures & 3 \\
\hspace{3mm}Excluded: unsuitable for SDM coding & 1 \\
\hline
Development consultations included in analysis & \textbf{7} \\
\hline
Testing set (held out) & 15 \\
\hline
\end{tabular}
\end{table}

From total agreement counts and normalized agreement rates per OPTION12 item across all (3) models,
Table \ref{tab:total_correct_per_item_normalized},
we can see that OPTION Item 7 has the highest agreement (42.9\%, 9 counts), while OPTION Items 1 \& 8 are the hardest (0\%) for all AI models.






\begin{table*}[t]
\centering
\small
\begin{tabular}{cccc}
\toprule
\textbf{OPTION12 item / summary} & \textbf{Gemma3:12b (\%)} & \textbf{Llama3.1:8b (\%)} & \textbf{Mistral7b (\%)} \\
\midrule
1  & \textbf{0}    & \textbf{0}    & \textbf{0} \\
2  & \textbf{42.9} & \textbf{42.9} & 28.6 \\
3  & \textbf{42.9} & 28.6 & 14.3 \\
4  & 28.6 & \textbf{57.1} & 28.6 \\
5  & 14.3 & \textbf{28.6} & \textbf{28.6} \\
6  & \textbf{42.9} & 14.3 & 0 \\
7  & 57.1 & 0    & \textbf{71.4} \\
8  & \textbf{0}    & \textbf{0}    & \textbf{0} \\
9  & \textbf{14.3} & 0    & 0 \\
10 & \textbf{42.9} & 0    & 14.3 \\
11 & 0    & \textbf{14.3} & \textbf{14.3} \\
12 & 0    & \textbf{57.1} & 14.3 \\
\midrule
{Total consensus count (over all items across 7 files)} & \textbf{20} & 17 & 15 \\
{Mean consensus per file (out of 12)} & \textbf{2.86} & 2.43 & 2.14 \\
{Mean consensus per file (normalized, \%)} & \textbf{23.83} & 20.25 & 17.83 \\
\bottomrule
\end{tabular}
\caption{Model-wise agreement per OPTION12 item between LLMs and human annotators across 7 files, with overall summary statistics. Bold indicates the best-performing model(s) for each row.}
\label{tab:agreement_per_item_summary_per_model}
\end{table*}

\begin{table}[t]
\centering
\tiny 
\begin{tabular}{ccc}
\toprule
\textbf{OPTION12 Item} & \textbf{Total Correct Count} & \textbf{Agreement (\%)} \\
\midrule
1  & 0 & 0.0 \\
2  & 8 & 38.1 \\
3  & 6 & 28.6 \\
4  & 8 & 38.1 \\
5  & 5 & 23.8 \\
6  & 4 & 19.0 \\
7  & 9 & 42.9 \\
8  & 0 & 0.0 \\
9  & 1 & 4.8 \\
10 & 4 & 19.0 \\
11 & 2 & 9.5 \\
12 & 5 & 23.8 \\
\bottomrule
\end{tabular}
\caption{Total agreement counts and normalized agreement rates per OPTION12 item across all models (3) and files (7), i.e., out of 21 possible agreements per item.}
\label{tab:total_correct_per_item_normalized}
\end{table}










The result from each model is shown in Table \ref{tab:agreement_per_item_summary_per_model}.
From Table \ref{tab:agreement_per_item_summary_per_model}, it shows that Gemma3:12b has the largest number count of consensus (20) with human annotation over 7 files/patients consultations, followed by Llama3.1:8b (17) and Mistral7b (15).
The mean consensus count per consultation (i.e., approximately how many items) is 2.86, 2.43, and 2.14 for these three models, which means there are, on average, between 2 and 3 items that LLMs agree with human annotators exactly, out of 12 items. 
The normalized percentage values are 23.83\%, 20.25\%, and 17.83\% for the three LLMs, which are very low, i.e., less than 25\% of chance they can be right individually.

From the individual model performance table (\ref{tab:agreement_per_item_summary_per_model}), we can also see that: 
1) The highest agreement on the OPTION item 7 is due to higher performances from Gemma3:12b (57.1\%) and Mistral7b (71.4\%), but Llama3.1:8b has 0\% correct on this item.
2) Three models perform very differently across 12 items, each model having 0\% scores on some items but not always the same items.
3) OPTION items 2 to 5 never have 0\% scores. 
These findings further verified our methodology design on how to use model augmentation (e.g., multiple models) to achieve better performance across all OPTION items, using a collective effort of LLMs.




In summary, Table \ref{tab:agreement_per_item_summary_per_model} and \ref{tab:total_correct_per_item_normalized} answered our \textbf{RQs} 1a): OS-sLLMs are far from human performances on overall OPTION 12 items except for specific Items, e.g., Item-7 (Mistral7b achieving a promising score 71.4\%);  1c) OPTION items 1 and 7 are extremely hard for OS-sLLMs, and other Items have different difficulty levels across models.
1d) Medical domain OS-sLLMs totally failed, in comparison to general domain ones.

Some further findings from these two tables on OPTION12 items' difficulty level are: 
1) 
Easy: 
Item 7 with 42.9\%; 
2) Medium:
Items 2, 4, with 38.1\%;
3) Hard:
Items 1, 8, with 0\%.
The variation in performance across OPTION12 items suggests that different cognitive and discourse-analysis skills are required for successful coding. For example, Item 1 ("The clinician draws attention to an identified problem as one that requires a decision-making process") requires the model to perform temporal framing and discourse-structure reasoning. The model must identify not only whether a decision is discussed, but also whether the clinician explicitly frames the situation as requiring a decision at the appropriate stage of the consultation. Several models recognized later treatment discussions but failed to distinguish them from the initial decision-framing process.

In contrast, Item 8 ("The clinician checks that the patient has understood the information") requires interactional reasoning and clinician-verification recognition. Models must correctly identify conversational acts in which the clinician actively assesses patient understanding, distinguish them from general information provision, and avoid confusing statements made by patients or caregivers with clinician verification behaviours. The low performance on this item suggests that current OS-sLLMs struggle with accurately interpreting conversational roles and communicative intentions within clinical dialogues.

In addition to the measurements based on the counting of exact matching score points between LLMs and human, we also explore the sequential correlation between LLM scores and human scores over OPTION12 items. 
To do so, we further evaluate the alignment between LLM-generated scores and human annotations using Pearson ($r$) and Spearman ($\rho$) correlation coefficients across OPTION12 items. As shown in Table~\ref{tab:correlation_results}, Gemma3:12b achieves the highest correlation with human annotations ($r=0.513$, $\rho=0.587$), outperforming Llama3.1:8b and Mistral7b. This also shows the moderate correlation level between LLMs and humans, which can be promising for future development on this direction.

\begin{table}[t]
\centering
\small
\begin{tabular}{lcc}
\toprule
\textbf{Model} & \textbf{Pearson ($r$)} & \textbf{Spearman ($\rho$)} \\
\midrule
Gemma3:12b & \textbf{0.513} & \textbf{0.587} \\
Llama3.1:8b & 0.180 & 0.045 \\
Mistral7b & 0.180 & 0.101 \\
\bottomrule
\end{tabular}
\caption{Correlation between LLM scores and human annotations across OPTION12 items over 7 patient files.}
\label{tab:correlation_results}
\end{table}

Correlation measures whether models preserve relative scoring trends across OPTION12 items, whereas exact agreement requires predicting the precise 0–4 score assigned by human annotators. Therefore, moderate correlations alongside low exact agreement suggest that the models capture broad SDM patterns but remain insufficiently reliable for fully automated coding.

\section{Error Analysis from Pilot Study}

\subsection{Data Preprocessing}
From the development set, we also find a data noise issue from human evaluation. 
For instance, the Caregivers can be CG but also CG1 or CG2 when there are multiple caregivers in the consultation. 
The initial preprocessing of our method did not take this into account, so the text following CG1 was not extracted for LLM use in the prompting stage. 
Our thought is that for OPTION12 items for SDM measurement, it concerns more what doctors and patients say, instead of caregivers.
For such files, we plan to \textbf{investigate} \textit{whether CG input/text has a significant impact on the SDM}. So, we will relocate these files (with corrected preprocessing and extraction) to the testing phase and remove them from example-based training (to prevent the model from seeing them before testing) to see whether the SDM scores differ substantially. 
{There are 2 such files with CG and CG1, ID 110022/23, indicating two caregivers in the consultations.
}

There are also three files with different timing formats, e.g., having the timing rounded with `[]' while other files do not. Such files were not correctly preprocessed, so the LLMs had no output. 
We relocated 3 of such files from development to the test phase.

\subsection{LLM Qualitative Error Categorization}




In many cases, LLMs can be correct; however, we list the situations when they go wrong with categorization, as in Table \ref{tab:longtable-qualitative-errors}. We hope this can help future researchers tackle some of the challenges. This list of categories partially reflects our \textbf{RQs}: \textbf{1a}: whether OS-LLMs perform well, and \textbf{1b}: whether OS-LLMs give different views.

The categorized LLM-related issues with examples are 1) inability on temporality-management meaning that it matters for the situation on if the information is given/provided in the beginning, middle, or end of the consultation; 
2) mis-interpreting the item/option, e.g. understanding \textit{what it is} instead of \textit{a decision is needed} or \textit{what are the options};
3) giving right scores as human annotators but wrong or inaccurate evidence or justification;
4) justification or evidence are right but scores are different from human annotators;
5) missing justification or evidence;
6) mis-understanding/swapping role-play;
and 7) made-up/hallucinated evidence.








\subsection{Medical vs General LLMs}
Why do medical models fail?
This is an interesting finding, becasue 
MedLlama and 
Meditron
perform worse than
Gemma,
Llama, and
Mistral.
One possible explanation is that:
OPTION12 coding is not primarily medical knowledge.
It is:
discourse analysis,
conversational reasoning,
and interaction assessment.
General LLMs may be better trained on these abilities.

\section{Conclusions and Future Work}




This study presents the first investigation of open-source smaller language models (OS-sLLMs) for automated assessment of shared decision making using the OPTION12 framework in Dutch melanoma consultations. Using a development set of human-annotated consultations, we evaluated three general-domain models (Gemma3:12b, Llama3.1:8b, and Mistral7b) and two medical-domain models (MedLlama2:7b and Meditron7b).
Our findings show that general-domain models consistently outperformed the tested medical-domain models. While the medical models frequently generated hallucinated outputs and exhibited difficulties following task instructions, the general-domain models demonstrated moderate alignment with human annotations. Among them, Gemma3:12b achieved the strongest overall agreement, obtaining the highest correlation with human scores (Pearson = 0.51, Spearman = 0.59). Item-level analysis revealed substantial variation in performance across OPTION12 categories, suggesting that different aspects of shared decision making pose different levels of difficulty for current LLMs.

Beyond quantitative evaluation, we identified several recurring categories of LLM errors, including failures in temporal reasoning, role attribution, interpretation of decision-making options, evidence selection, and hallucinated justifications. These findings provide insights into the limitations of current OS-sLLMs and highlight specific challenges that must be addressed before reliable deployment in clinical quality assessment workflows.

Overall, our results suggest that open-source smaller language models are not yet capable of replacing human annotators for OPTION12 coding. However, they show potential as supportive tools within human-in-the-loop annotation pipelines, particularly for accelerating coding and assisting quality assurance while preserving privacy through local deployment.






Several directions remain for future investigation. First, the optimized prompting strategies and model configurations identified during the development phase will be deployed on the held-out test set to evaluate the generalizability of the findings. This evaluation will also assess the effectiveness of the proposed Judge-LLM framework for resolving disagreements among multiple OS-sLLMs.
Second, future work will investigate ensemble and consensus-based approaches that combine the complementary strengths of different models across OPTION12 items. The observed variability in item-level performance suggests that model collaboration may provide more reliable assessments than any individual model alone.
Third, additional work is needed to improve medical-domain OS-sLLMs. Despite their domain specialization, the tested medical models exhibited substantial hallucination and instruction-following issues. Future experiments will examine whether prompt optimization, supervised fine-tuning, parameter-efficient adaptation methods (e.g., LoRA), or domain-specific instruction tuning can improve their reliability for SDM assessment tasks.
Fourth, we plan to quantify inter-rater reliability between human annotators and compare LLM performance against human-human agreement levels. Such analysis will provide a more meaningful benchmark for evaluating the practical usefulness of LLM-assisted coding.
Finally, future research will explore human-in-the-loop workflows in which LLMs provide preliminary OPTION12 annotations and evidence extraction while human experts verify and refine the outputs. This approach may reduce annotation burden while maintaining the transparency and reliability required in clinical research settings.

\clearpage
\onecolumn

\setlength{\LTleft}{0pt}
\setlength{\LTright}{0pt}

\begin{longtable}{p{0.18\textwidth} p{0.18\textwidth} p{0.32\textwidth} p{0.18\textwidth} p{0.08\textwidth}}
\caption{Qualitative analysis of LLM performance in shared decision making.} 
\label{tab:longtable-qualitative-errors}
\\
\toprule
\textbf{Issue Type} & \textbf{Item} & \textbf{LLM Interpretation} & \textbf{Human Annotation} & \textbf{Consensus} \\
\midrule
\endfirsthead

\toprule
\textbf{Issue Type} & \textbf{Item} & \textbf{LLM Interpretation} & \textbf{Human Annotation} & \textbf{Consensus} \\
\midrule
\endhead

\midrule
\multicolumn{5}{r}{Continued on next page} \\
\endfoot

\bottomrule
\endlastfoot

Temporality-management
& Item 1: The clinician draws attention to an identified problem as one that requires a decision making process
& LLM identifies the decision-making process at the end of the consultation rather than at the beginning. For example, Gemma and Mistral use the clinician's statement about discussing treatment as evidence, even though the decision framing occurs too late in the consultation.
& Human-1: 1; Human-2: 2
& 0 \\

\addlinespace

Misinterpreting the item/option
& Item 2: The clinician states that there is more than one way to deal with the identified problem
& Gemma cites explanation of melanoma and Breslow thickness as if it were discussion of multiple management options, but the quoted evidence does not present alternatives.
& Human-1: 1; Human-2: 1
& 1 \\

\addlinespace

Misinterpreting the item/option (continued)
& Item 1
& Llama 3.1 treats the clinician's check of what was previously explained about melanoma as if it were introducing a decision process, while it is actually background clarification.
& Human-1: 1; Human-2: 2
& 0 \\

\addlinespace

Scoring right but justification/evidence not accurate
& Item 3: The clinician assesses the patient’s preferred approach to receiving information to assist decision making
& Mistral assigns a low score appropriately, but the justification and evidence focus on explanation of Breslow thickness rather than the patient's preferred way of receiving information.
& Human-1: 1; Human-2: 1
& 1 \\

\addlinespace

Scoring right but evidence not accurate
& Item 4: Clinician lists all options, including doing nothing
& Mistral's justification correctly notes that treatment options were not explicitly discussed, but the selected evidence sentence is unrelated and appears arbitrary.
& Human annotation: no listing; not detailed about wait-and-see
& 1 \\

\addlinespace

Justification / evidence right but score differs from human
& Item 1 (File 130003)
& Gemma, Llama, and Mistral correctly identify that the clinician introduces the treatment discussion, but they over-score it as full decision framing or shared decision making.
& Human: 1, 1, 1
& 1 \\

\addlinespace

Missing justification/evidence
& Item 3 (File 130003)
& Gemma assigns score 1 and states that there is no explicit assessment of the patient's preferred learning style, but provides no actual evidence excerpt.
& Human: 1, 1, 1
& 1 \\

\addlinespace

Role misunderstanding
& Item 8: The clinician checks that the patient has understood the information
& Mistral uses a caregiver utterance as if it were spoken by the clinician, leading to an incorrect attribution of understanding-check behavior.
& Human: 0, 0, 0
& 0 \\

\addlinespace

Hallucinated evidence
& General issue
& In some cases, the LLM produces evidence or statements that are not grounded in the transcript at all.
& --
& -- \\

\addlinespace

Evidence / justification partly right but score too high
& Item 1 / later example from File 130003
& The clinician asks for agreement with surgery, and the evidence is relevant, but the model over-interprets this as balanced shared decision making, whereas the interaction is more directive.
& Human: 2, 2, 2
& 2 \\

\addlinespace

Evidence / justification okay but score too high
& Continued example
& Llama scores the clinician's brief acknowledgment of patient anxiety too highly, although the concern is only superficially explored.
& Human: 1, 1, 1
& 1 \\

\end{longtable}

\clearpage
\twocolumn

\section{Acknowledgment}
Funded by the European Union under Horizon Europe Work Programme 101057332. Views and opinions expressed are however those of the author(s) only and do not necessarily reflect those of the European Union or the European Health and Digital Executive Agency (HaDEA). Neither the European Union nor the granting authority can be held responsible for them.
 The UK team are funded under the Innovate UK Horizon Europe Guarantee Programme, UKRI Reference Number: 10041120.




 


\appendix

\section{Data Statistics vs SDM}

\begin{table*}[t]
\centering
\small
\begin{tabular}{llrrrrrrc}
\toprule
\textbf{File} & \textbf{Speaker} & \textbf{Turns} & \textbf{Duration} & \textbf{Ratio (\%)} & \textbf{Text Len.} & \textbf{Total Dur.} & \textbf{Segments} & \textbf{Human SDM Mean} \\
\midrule
110019 & Doctor    & 104 & 13m 38s & 49.9 & 14156 & 27m 18s & 272 & 0.83 \\
110019 & Patient   & 97  & 7m 01s  & 25.7 & 3777  & 27m 18s & 272 & 0.83 \\
110019 & Caregiver & 71  & 6m 39s  & 24.4 & 5168  & 27m 18s & 272 & 0.83 \\
\midrule
110022 & Doctor    & 70 & 11m 02s & 70.5 & 8091 & 15m 39s & 155 & 1.42 \\
110022 & Patient   & 61 & 3m 46s  & 24.1 & 2086 & 15m 39s & 155 & 1.42 \\
110022 & Caregiver & 24 & 0m 51s  & 5.4  & 405  & 15m 39s & 155 & 1.42 \\
\midrule
130002 & Doctor    & 87 & 20m 23s & 70.6 & 19461 & 28m 52s & 204 & 0.67 \\
130002 & Patient   & 41 & 3m 10s  & 11.0 & 2053  & 28m 52s & 204 & 0.67 \\
130002 & Caregiver & 76 & 5m 19s  & 18.4 & 3646  & 28m 52s & 204 & 0.67 \\
\midrule
130003 & Doctor    & 79 & 14m 09s & 67.5 & 13757 & 20m 58s & 176 & 1.00 \\
130003 & Patient   & 67 & 4m 59s  & 23.8 & 4054  & 20m 58s & 176 & 1.00 \\
130003 & Caregiver & 30 & 1m 50s  & 8.7  & 1390  & 20m 58s & 176 & 1.00 \\
\midrule
130004 & Doctor    & 35 & 8m 40s & 80.8 & 7759 & 10m 44s & 73 & 1.42 \\
130004 & Patient   & 20 & 0m 50s & 7.8  & 427  & 10m 44s & 73 & 1.42 \\
130004 & Caregiver & 18 & 1m 14s & 11.5 & 502  & 10m 44s & 73 & 1.42 \\
\midrule
130005 & Doctor  & 86 & 10m 17s & 62.5 & 11496 & 16m 28s & 173 & 1.17 \\
130005 & Patient & 87 & 6m 11s  & 37.5 & 4147  & 16m 28s & 173 & 1.17 \\
\midrule
130006 & Doctor  & 45 & 9m 13s & 69.1 & 6973 & 13m 20s & 86 & 0.42 \\
130006 & Patient & 41 & 4m 07s & 30.9 & 1202 & 13m 20s & 86 & 0.42 \\
\bottomrule
\end{tabular}
\caption{Speaker-level descriptive statistics with corresponding human-annotated SDM mean scores (OPTION12) for each consultation.}
\label{tab:speaker_statistics_with_sdm}
\end{table*}

We are curious about whether the speaker frequency, duration, and ratio from (doctor, patient, caregiver) matters regarding shared decision making outcomes.
To investigate this aspect, we display the statistics of such.


Table~
\ref{tab:speaker_statistics_with_sdm} presents speaker-level descriptive statistics, including speaking turns, speaking duration, duration ratio, text length, total consultation duration, and total number of segments for the seven development-set consultations.

Looking into the last column on the right-hand side of Human SDM Mean, as well as the statistics from this table, the higher SDM scores are achieved by patient files 110022 and 130004 (SDM both 1.42), and 130005 (SDM 1.17). 
They have the doctor speaking ratio (70.5, 80.8, 62.5), which is a wide range. The other files with lower SDM scores have doctor speaking ratio values falling in the range [62.5, 80.8], so we can not make a conclusion on the impact of speaker turns on SDM scores.


\bibliography{custom}


\end{document}